\documentclass{superfri}
\usepackage[hidelinks]{hyperref}
\usepackage{todonotes}

\begin{document}

\author{Edmon Begoli\footnote{\label{ornl} Oak Ridge National Laboratory, Oak Ridge, Tennessee, USA}, Jim Brase\footnote{\label{llnl} Lawrence Livermore National Laboratory, Livermore, California, USA}, Bambi DeLaRosa$^3$, Penelope Jones$^3$, Dimitri Kusnezov\footnote{\label{doe} US Department of Energy, Washington DC, USA}$^,$\footnote{ Keynote Address, Supercomputing Frontiers Europe 2018, Warsaw. The views expressed in the article do not necessarily represent the view of the U.S. Department of Energy or the United States.}, Jason Paragas$^2$, Rick Stevens\footnote{\label{anl} Argonne National Laboratory, Argonne, Illinois, USA }, Fred Streitz$^2$, Georgia Tourassi$^1$} 

\title{Precision Medicine as an Accelerator for Next Generation Cognitive Supercomputing}

\maketitle{}

\begin{abstract}%
In the past several years, we have taken advantage of a number of opportunities to advance the intersection of next generation high-performance computing AI and big data technologies through partnerships in precision medicine. Today we are in the throes of piecing together what is likely the most unique convergence of medical data and computer technologies. But more deeply, we observe that the traditional paradigm of computer simulation and prediction needs fundamental revision. This is the time for a number of reasons. We will review what the drivers are, why now, how this has been approached over the past several years, and where we are heading.

\end{abstract}

\section{Computing and Prediction}
\label{sec:intro}
Prediction is part of our everyday lives. From early recorded predictions with deep societal impact, such as the prediction of a solar eclipse by Thales of Miletus in 585 BC, to our daily use of prediction in weather, tides, sports or science, the uses have become ubiquitous---although the consequences of poor predictions are not all the same. Today we are turning to science-based prediction to inform increasingly important problems. These are typically situations which cannot be tested experimentally, from emergency preparedness or response, to safety and security in its many dimensions. We have to protect against catastrophic events, things that cannot be instrumented or measured beforehand, but that require enough understanding to invest limited resources optimally. Our missions are the drivers of our technology needs. They include protecting the US energy sector from man-made or natural disasters and ensuring that our nuclear deterrent will respond as needed. Increasingly, though, we are turning to supercomputing as a tool that enable us to answer a new and growing set of urgent questions. As we have developed our tools to meet this class of problems, we find that, in addition to our direct responsibilities, our computing capabilities have led to us becoming, in recent years, a ``go-to" agency
for informing urgent decisions. We are involved in a regular cadence of events, including responses to Fukushima-Daiichi, the Gulf oil spill, Ebola, the Christmas day ``underwear'' bomber, aviation security and so forth. Consequently, the certainty, or, importantly, the quantified uncertainty, that we can provide with any set of complex predictions underpins how we develop our next generation hardware and software tools.\footnote{This manuscript has been in part authored by UT-Battelle, LLC, under contract DE-AC05-00OR22725 with the U.S. Department of Energy. The United States Government retains and the publisher, by accepting the article for publication, acknowledges that the United States Government retains a non- exclusive, paid-up, irrevocable, world-wide license to publish or reproduce the published form of this manuscript, or allow others to do so, for United States Government purposes. }

Our tools for prediction are built around supercomputing simulations. Simulations can be separated into two broad classes or approaches \cite{twoclass} which impact how we think about computer architecture development. In the first and more conventional approach, the model is defined and driven by a  domain expert. Problems are well-defined and well-posed. Usually a mathematical theory  exists, a model can be stated explicitly, and much of our effort is spent on refining the numerical precision of the solution. Researchers scale such ``single-physics'' calculations to the maximum scale the available computer can handle. The problems start with known degrees of freedom, perhaps defined in a phase space or a Hilbert space, there are observables to measure, and the researcher usually already lives in a discretized space, spending time controlling the accuracy of the numerical measurements under investigation. It might be computing observables in lattice QCD, or free-energy surfaces, or protein folding.

The second class are those which we are applying to high-leverage decisions; where the promise of supercomputing is driving us to address complex, multifaceted issues of national importance, but where the approach to prediction still requires significant mathematical development. These questions are not necessarily well-posed and, in many cases, cannot be clearly stated in mathematical terms. They are often time-urgent, typically not asked by scientists but arising from events where science needs to have a voice. There is usually no good starting point---no Hamiltonian or Lagrangian, no observables to turn to, no well-defined phase space to start an analysis. Yet these are the classes of questions that people care about. These are complex, multi-scale, multi-disciplinary problems and can seldom be addressed by a single domain expert but rather require multidisciplinary teams of experts. Discovery also lies here, where we are limited by our imagination and the risk we are willing to take.

It is this second class of problems that we will focus on here.  In particular what prediction means in this emerging problem domain and what it might take to strengthen our prediction capabilities. When we are faced with critical events or crises, today we are likely to ask: Can simulation play a role? This is not always the case, but if the answer is yes (noting that not everything is suited to be scientifically-informed) we encounter immediate questions such as:

\begin{itemize}
\item What does it mean?
\item What actions are needed and when? 
\item What is your confidence? 
\item What are the risks? 
\item What happened?
\item Can it happen again?
\end{itemize} 

\noindent We often do not know if we are asking the right questions or even the right people, or whether we are positioned to answer them. Nevertheless, increasingly we are challenged to answer questions such as these that drive our decisions and providing the best means to make them is the problem at hand. In our work with this class of problems, we have built a discipline of capturing our best mathematical sense of the overall uncertainty associated with the predictions, because there are consequences to being wrong. We term this disciplined approach Uncertainty Quantification (UQ). 
It includes the ability to numerically verify the computational tools, to understand the dependence of predictions on model uncertainties, and to experimentally validate predictions against available relevant or adjacent data. There are many successes to date, and certainly significant money saved in deferred costs. In all these cases, there is typically a cost for inaction as well.

\section{Uncertainty Quantification is likely NP Hard}

\subsection{Limitations of Our Current Approach}

In our 70 year-old approach to modeling physical systems, designed from the paradigm of von Neumann architectures, UQ has developed as a mathematical afterthought. We usually start in a space of discretized guiding equations and meshes, where a prescribed set of instructions defines the time evolution of the problem at hand from initial to final time. We then add in a post-hoc manner tools to accrue and capture overall uncertainty. These could be Bayesian approaches, Monte Carlo or Latin hypercube sampling (LHS), running ensembles of calculations with different models/parameters, and so forth. 
However, at the time of their introduction, these methods did not anticipate the approaching richness of the numerics or experiments. This is illustrated in~\figref{Cog_engine} (a) where a UQ approach creates an overall cloud of predictions consistent with the ranges of uncertainties in the model-based computational approach.

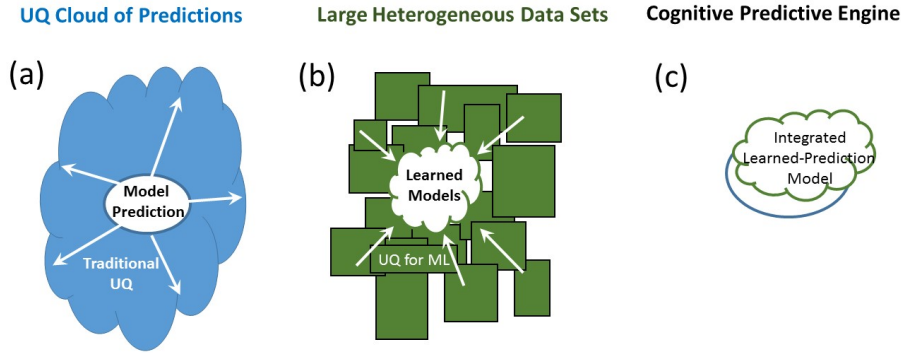
\fig{width=12cm}{Cog_engine}{(a) Our current approach to UQ. Models make certain predictions after which we vary assumptions, approaches, initial conditions etc to better capture the full cloud of predictions. (b) In Machine Learning (ML) approaches, reduced dimensional manifolds are extracted from higher dimensional data sets, with data incompleteness and noise, and learned models extracted. This captures our post-Moore's Law activities today. Here `UQ for ML' needs to be developed to inform uncertainties in learned models from the overall data. (c) The path we are taking to converge AI/ML based learned models with model based predictions so that all numerical and experimental data can help self-inform predictions, effectively building in UQ into a ``cognitive predictive engine''.}

We have led in the development of rigor in this field over the years, but today we argue that the approach is dated and needs fundamental revision. Our historic approach is a systematic one that involves comparisons between numerical simulations and experiments ranging in scale all the way from full systems, down through subsystems, benchmark scales and unit scale problems\cite{vandv}. Data is used to test all available length-scales of the simulations. When the richness of the simulations exceeds that of the experiments, such `point-wise' comparisons define a reasonable approach. Today, with the increasing richness of data, already at peta-byte scales, we are plagued with the challenge of trying to validate both spatial and temporal data against simulation data also at peta-byte scales. The technology trajectory in sensors and detectors (leading to exquisite measurements over many length and time scales) coupled with the advent of exascale computations will bring both numerical and experimental data to exascales. 
Existing approaches to UQ and accompanying computer architectures are not prepared for this. As we explore alternatives, we will need to retain pragmatic solutions since we have missions to deliver; we need mathematical frameworks that are implementable in practice. 

To further add to the complexity, the extraction of physical models from data alone, in either classical or quantum mechanical systems, is known to be NP-hard (see for example \cite{cubitt2012extracting}). It is likely the case that this approach to uncertainty quantification will fall into the same complexity class. Herein lies the challenge---it is possible that our current approach to UQ for multi-scale simulations is in fact intractable on today's (von Neumann-based) technology path. 

\subsection{A Path Forward}

Over the past three or so years, we have been examining how to make progress, building on the advances in artificial intelligence (AI) and machine learning (ML) in particular. We know from the study of games, such as Go, Chess or Poker, that heuristics can go quite far in pushing beyond what traditional computation limits might have suggested. Machine Learning (ML) methods seek to generate reduced dimensional manifolds from higher-dimensional data, generating effective models from the data. This is illustrated in~\figref{Cog_engine} (b) where large data sets are aggregated, often incomplete or noisy, and learned models are extracted. To provide a sense of the uncertainty in the learned models created through this process, UQ for ML  approaches need to be developed. Contrast this to our traditional HPC approach to simulation which takes physical models, captured in discretized forms, and provides a numerical cloud of predictions for given problems. We are seeking to approach the (potentially) NP hard nature of UQ by finding a means to connect these two worlds, where the ML-learned models are guided by the physical models, perhaps in the way heuristics can be used to guide solutions to NP hard problems. We have been approaching this by trying to bring together hardware and software technologies in areas as illustrated in~\figref{convergence1}. 

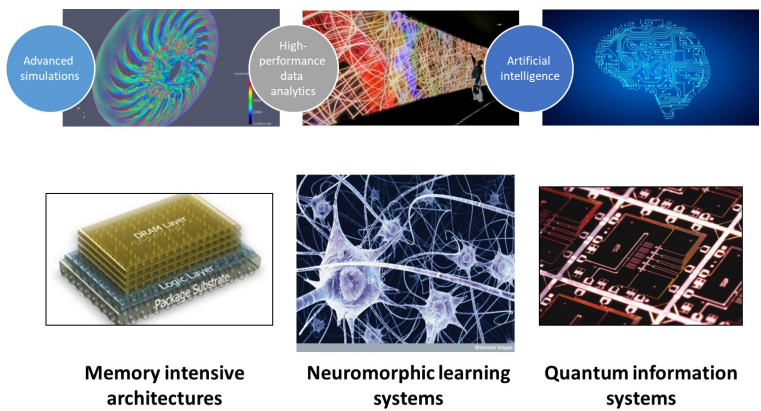
\fig{width=10cm}{convergence1}{(Top) We are driving hardware and software convergence across three fields. (Bottom) Some of the specific artificial intelligence (AI) class technologies we are exploring.}

The challenges lie well beyond just hardware and software frameworks. Many machine learning methods, such as deep neural networks, are not amenable to analysis of the propagation of uncertainty. While training performance can be used to bracket effectiveness of an approach, there is no current mathematical basis for UQ in ML that we can turn to. Consequently converging AI, Big Data and HPC poses interesting mathematical challenges in terms of providing quantifiable uncertainty in predictions. We are faced with the following dilemmas: There is too much data to understand, let alone compare in its full richness to simulation today. UQ approaches are based on our traditional computational paradigms, but in the age of ML and AI it has no firm mathematical foundation. Extraction of governing dynamical equations from any amount of data is provably NP-hard \cite{cubitt2012extracting}. Breaking away from the model of how we currently approach predictive simulation on today's computer architectures will require a leap of faith that is hard to do for those heavily invested in our legacy tools, solvers and methods. However, breaking away from our current model is likely the only way we will be able to effectively meet the demands of the future. This is illustrated in~\figref{Cog_engine}(c) where we are seeking both hardware and software means to create the union of data and model informed predictions simultaneously so that UQ can be built into a new approach to predictive simulation. Enter precision medicine. 

\section{Why Precision Medicine? }

Precision medicine\footnote{By precision medicine, we use the definition of Yamamoto: ``Use a massive data network that aggregates and analyzes information from huge patient cohorts, healthy populations, experimental organisms – and reaches toward disease  mechanisms, and precision diagnosis and treatment for each individual." } is an innovative approach intended to provide insight and inform physicians on how to best tailor medical treatment plans, procedures, and therapies for individual patients by taking into account the complexity of their specific health profile. It is a computationally-intensive exercise enabled by the utilization of high performance computing capabilities and AI frameworks and the integration of large-scale, diverse, complex layers of health and disease-related data such as electronic health records (EHR), medical imaging, various -omics data (i.e. genomic, epigenomic, phenomics, proteomic, metabolomics), basic sciences research, systems biology models, molecular, pharmaceutical, and other biomedical related data \cite{gligorijevic2016integrative,hawgood2015precision}, illustrated in~\figref{surpassing}. 
With AI, data generated from very large patient populations may be `learned' into models that are then applied to the individual patient or patient cohort, allowing prediction of what treatment, therapy, or prevention strategy will work best. To make precision medicine work today, medical fields rely on methods from statistics and computer science---mainly statistical models, machine learning, and data mining. These statistics and machine learning models need comprehensive data products (-omics-based, clinical history, laboratory results, lifestyle, environmental, etc.) which needs to be engineered with consistency, precision, and quality. 

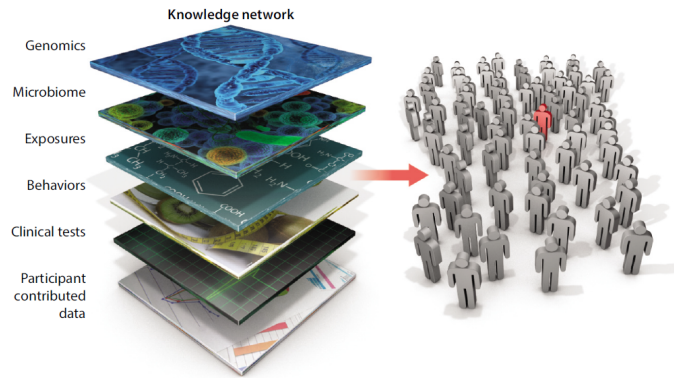
\fig{width=10cm}{surpassing}{Precision medicine aims to integrate large-scale, diverse datasets reflecting the inherent complexity of biological systems and their role in health or underlying disease.}

We are challenged currently with a mistiming: the pressing questions that we have outlined above have not yet reached a level where paradigm change is unavoidable.  However, industry today is making technology choices that are starting to define their architectural directions with respect to  ML and AI and the opportunities exist today for aligning these with the direction we would like predictive simulation to evolve. To start to explore a transition in predictive simulations from today's paradigm to a possible future one, we have turned to a field that is replete with data but not burdened by how we historically developed complex multi-scale predictive simulations. Our strategy to attack this challenge is to seek large force multipliers by: 
\begin{itemize}
\item Finding problems that can serve as attractors for new ideas, as a strategy to draw in broader thinking and resources;
\item Forcing the rethinking of traditional paradigms by challenging researchers with qualitatively new classes of prediction through a richness of data;
\item Increasing the number of people teaming together to find common solutions;
\item Using the qualities of data to change how we think of many of our traditional approaches from post Moore's Law hybrid architectures to uncertainty quantification to codes. 
\end{itemize}

\noindent Precision medicine provides a means to meet these goals. We have started by building around three main elements: (a) {\bf Technology}: Development and testing of next generation hardware and software technologies; (b) {\bf Data}: Aggregation of new sources of complex data, curated by experts and restructured for large, high-end, scalable  architectures; and (c) {\bf Partnerships}: Buying down risk through engaged partners who are driven by impacting outcomes in technology or precision medicine. This includes other agencies, private sector, international partners and agencies, and academia.  In the next few sections we will outline some of these, with the overall intent of informing our technology convergence challenge.

\subsection{Joint Development of Advanced Computing Solutions for Cancer ``JDACS4C"}

In mid-2015 we launched a partnership with the National Cancer Institute (NCI) that has five efforts that define it: i. molecular level understanding, ii. patient health trajectories, iii. population-level studies, and two crosscutting efforts that include UQ and a cancer distributed learning environment project called CANDLE. These projects involve four of our national laboratories (Argonne, Livermore, Los Alamos and Oak Ridge), the NCI's Frederick National Laboratory, as well as NCI scientists. Data sources are provided by the NCI, and we focus on using the NCI challenges to develop new technologies which can be used to advance both agency's missions. 

At the molecular level, we are examining the biology of oncogenic RAS proteins in lipid membranes. RAS-based cancers are the most intractable challenge in oncology today, defeating our best technologies. Mutated RAS is responsible for 93\% of all pancreatic cancers, 42\% of all colorectal cancers, 33\% of all lung cancers, with over 1 million deaths worldwide and no effective inhibitors. The mechanisms of RAS activation are not well understood, hindering the discovery of effective therapeutics. We are developing a hybrid AI/HPC molecular-scale methodology for RAS activation to help model development of targeted therapies for RAS-based cancers. Our approach is highlighted in \figref{pilot2}. 
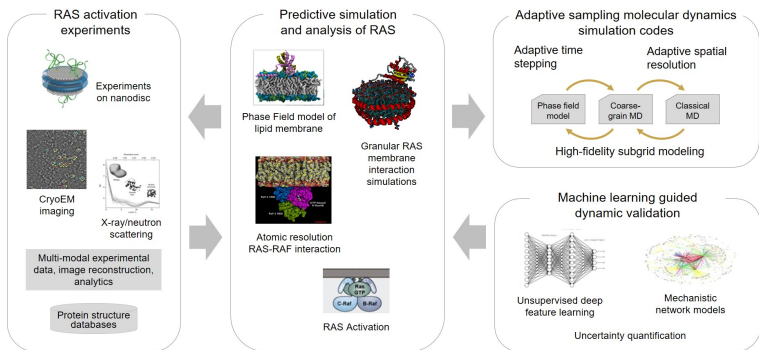
\fig{width=10cm}{pilot2}{The schematic approach of the RAS-based cancer effort is shown here.} 
New validation experiments coupled to atomistic and coarse-grained molecular dynamics simulations are being used to dynamically model the behavior of RAS complex in the context of a complex membrane. Experiments incorporate advanced imaging and interaction data from such sources as cryo-EM, NMR, crystallography and neutron sources will focus on membrane-bound RAS, mutated RAS and RAS complexes will be used in conjunction with simulation to build accurate interaction parameters that will in turn power hyper-scaled simulation that access, for the first time, biologically relevant length and time scales for this problem. To date, we have demonstrated a novel multi-scale methodology linking atomistic and continuum resolution simulations, briefly captured in \figref{pilot2a}. This includes hyper parameter optimizations to characterize sub-states from molecular dynamics simulations, and machine learning on-the-fly during the running code to optimize parameter exploration. By scaling this capability to next-generation supercomputing technology, we aim to capture the behavior of RAS and mutated RAS molecules and complex interactions at the cellular membrane.

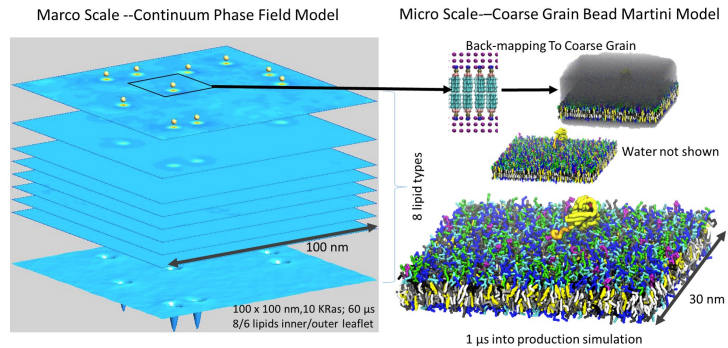
\fig{width=10cm}{pilot2a}{An example of macro to micro scale coupling with multiple lipid layers (left) at the cell membrane shown and RAS moving on the surface. Note the impacts are felt inside the cell. (right) Micro states are captured through hybrid AI methods that can draw in atomistic level dynamics and time scales.}

At the cellular level, we are developing AI as a means to improve treatments. Quantitatively predictive models for cancer therapy could one day support the treatment choices a physician and patient make toward achieving the best possible clinical outcome. To fulfill this vision, we need to be able to predict accurately the efficacy of cancer drugs in model systems.  To that end, we are building scalable models that integrate data derived from pre-clinical studies from cell lines, organoids, and patient-derived models (see ~\figref{pilot1_v2}). This framework accounts for and quantifies uncertainties in the state of systems of interest and takes advantage of information available in different forms and scales. Using machine learning we construct integrated models that include molecular characterization of the tumors, drug descriptors and drug fingerprints as well as information about dose and types of experiments.  The resulting family of learned models can predict drug response of tumors for standard of care drugs, drug combinations as well as drugs under development.  
Our approach outlined in~\figref{pilot1a}, addresses questions such as: Which molecular features or combination of features are most predictive of drug response?  How best to represent and quantify uncertainty in the predictions of the models?  Given the model prediction landscape that includes biological response and estimates of certainty what is the best course of action to validate the predictions and improve the models?  

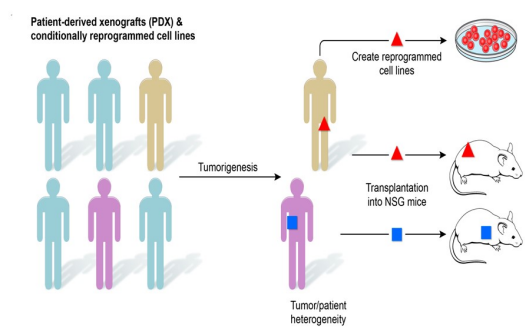
\fig{width=10cm}{pilot1_v2}{Patient Derived Xenograft Models: Tumors obtained from patients are grown in NSG mice to model drug response.}

We are constructing deep learning models from the response and feature sets of cell lines and PDX data that can be trained on integrated datasets from many experiments.  These training data include data from multiple labs, over a thousand cell lines, tens of thousands of compounds and thousands of drug combinations.  We are developing methods to interpret the resulting models to extract the most predictive features.  Existing training datasets are often unbalanced with some cancer types underrepresented and with response data often skewed.  To address these issues we are developing methods to augment our training data using generative adversarial networks (GANs).  GANs have been widely used in machine learning community and represent one of most innovative developments in representational learning of the past decade.  We are the first group to apply these methods to the Cancer drug response problem.  To address uncertainty we have adopted the "dropout" method of training our networks with dropout and then increasing the dropout fraction for inference.  This approach has been shown to be equivalent to a Bayesian approximation.   The resulting models when run in inference mode are effectively sampling from a distribution of models and as a result we can estimate the uncertainty in the predictions.  From this uncertainty estimate we can visualize the response landscape to gain some insight to where the model is relatively more or less confident.  Using this landscape we can identify needed experiments (or additional training data) that would reduce the uncertainty.

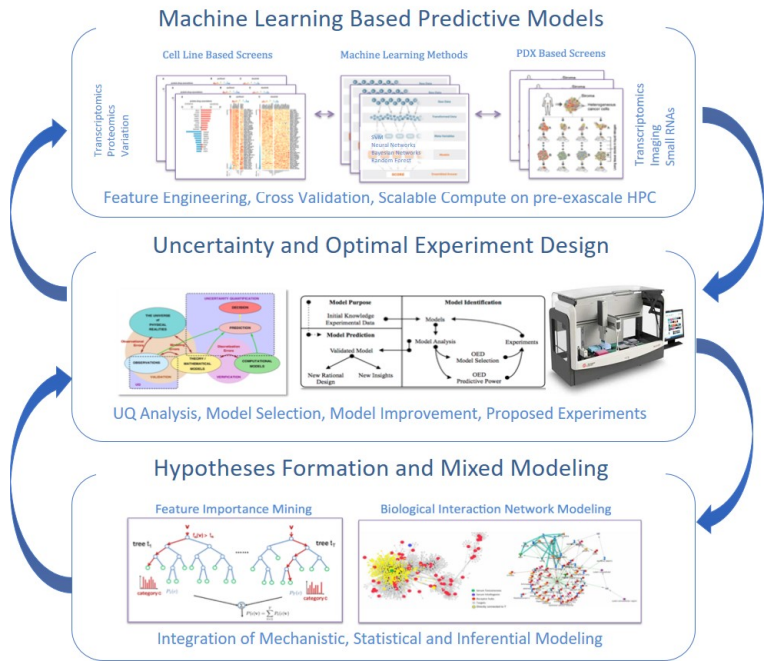
\fig{width=10cm}{pilot1a}{The schematic approach to the pre-clinical screening effort is shown here. }
 
At the population level, we are working to learn from individual cancer patients and build on the participation of cancer patients in understanding response to therapy, quality of life and outcomes at a national level. This project uses the existing SEER \cite{ries2003cancer} cancer registries and tumor registries as a hub for reporting therapeutic, genetic and genomic information across the United States. Building a strong, population-wide evidence base of cancer diagnosis, treatment, and outcomes information is critical for process improvement and modeling the cancer patient experience. Providing a hub for patients and providers to share information is critical to \textit{in silico} modeling of cancer at a population level.  A population-based understanding of cancer presentation, treatment and outcomes will in turn be critical to realizing the potential of cancer precision medicine. This project explores novel mechanisms for data sharing, consent, direct patient participation in research, and the impact of variations in care practices.  Increasingly available information on patient treatment response will be increasingly integrated providing insight critical into evaluating and understanding the effectiveness of cancer treatment choices. The approach is sketched in~\figref{pilot3}.

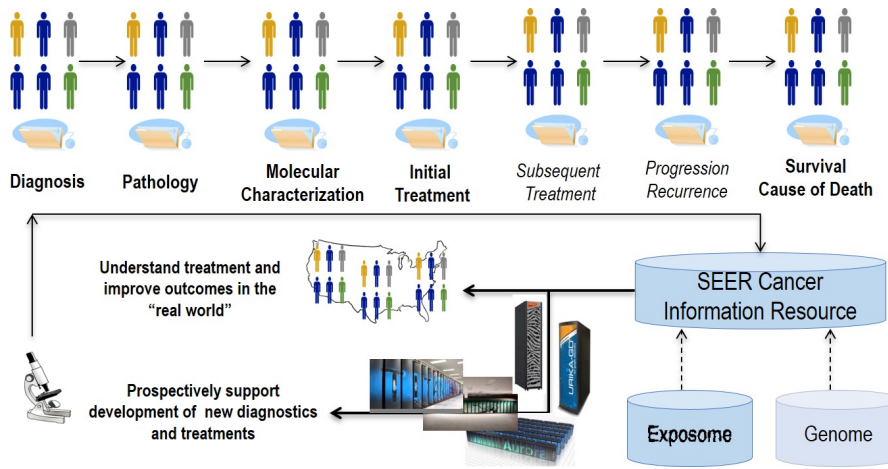
\fig{width=12cm}{pilot3}{The Surveillance Perspective: Improve the effectiveness of cancer treatment in the ``real world'' through computing}

We are developing algorithms and related informatics tools needed to automatically capture, integrate, analyze, and effectively utilize the information needed to support a comprehensive, scalable, and cost-effective national cancer surveillance program. Since critical amount of patient information is contained in unstructured clinical text such as pathology and radiology reports, initial efforts are focused on developing, benchmarking, and deploying across cancer registries advanced text comprehension algorithms to automatically and reliably abstract important patient information from cancer pathology reports \cite{gao2017hierarchical,yoon2016multi,qiu2018deep}. By collecting, linking, and analyzing additional heterogeneous data (such as pharmacy data, claims, biomarkers) we can generate a richer profile for the cancer patients, including healthcare delivery system parameters and continuity of care. These patient profiles will enable data-driven modeling and simulation patient-specific health trajectories laying foundation for in silico, large-scale evaluation of precision cancer therapies. Overall, these linked efforts provide the data infrastructure and methods to support precision oncology research at the population level.

Beyond these three scales of focus, there are two cross-cutting activities. For UQ, the computation required includes high-dimensional, non-convex, optimization, machine learning and deep learning, large-scale, long-time molecular dynamics, modeling strongly nonlinear stochastic systems, and carefully quantifying uncertainty in predictions. 
In addition, we are working in CANDLE to thoughtfully integrate machine learning methods with mechanistic modeling and to fully exploit the biological knowledge of molecular interaction networks in designing informative experiments. This exascale deep learning environment builds on existing open source deep learning frameworks through a software stack that includes workflow, scripting, execution engine and optimization. The effort pushes the envelope in rapid data integration methods that combine a variety of molecular level assays with imaging and phenotype assays to produce an integrated ``feature space'' that will underpin the basis for development of predictive models.

\subsection{AI and Precision Diagnostics -- Traumatic Brain Injury (TBI)}

Traumatic Brain Injury (TBI) is an area of clinical research perfectly poised in the second class of problems. The problem space for TBI is less well defined in that there are severe deficiencies in both the mechanistic understanding of TBI and currently crude clinical classification or diagnostic delineation of severity. Recently, understanding the cellular mechanisms of TBI has improved; however, given the broad classification of TBI diagnosis, this has limited translation into clinical relevance. The current TBI classification schema is dependent on a relatively insensitive and symptom-based approach \cite{manley2013traumatic}, preventing any mechanistic advancements to manifest into real-world application. Given that TBI is often disabling and an exponentially increasing source of morbidity and mortality in older adults \cite{haring2015traumatic}, improving understanding and developing mechanistic basis of treatment approaches would have a profound impact.

We are starting an effort to access, host and analyze a super-dataset by integrating several large and discordant datasets related to TBI. These data include a rich mixture of x-ray and MRI brain images, measurements of biochemical markers in the blood, and clinical data on symptoms and outcomes. Today, physicians have limited capability to bring these data together to provide the best possible diagnosis and treatment for individual patients. We believe cognitive computing could provide improved diagnosis and treatment. The Transforming Research and Clinical Knowledge in Traumatic Brain Injury (TRACK-TBI) dataset \cite{yue2013towards} is one of the most detailed ever collected in neuroscience and provides an opportunity to demonstrate the feasibility and utility of this approach. Additional public and private datasets are also being considered to expand the depth, richness, and complexity of the data. In the initial steps, using large-scale computing and machine learning, we are working on precision diagnostics for subjects in the TRACK-TBI study. These insights will be validated and verified by University of California, San Francisco medical school collaborators. Precision diagnosis represents a key frontier for making data-driven decisions in complex, uncertain, and discordant data environments. Longer term, the data analytic concepts developed here will work together with the other data sets outlined in this paper and could be important in the context of broader set of high-significance illnesses. By making TBI diagnosis more sensitive and accurate by reducing the time for MRI-based connectome analysis from a day to a few seconds creating a Real-Time Connectome (RTC). Initial efforts have demonstrated a factor of 1000 speedup in analysis. The tools are intended to provide physicians accessible tools to predict a TBI patient’s trajectory and optimize their treatment based on real-time analysis of clinical inputs, brain images, genomic data, and biomarkers. Already AI tools and massive parallelization for TBI data have initiated the development of precision TBI phenotypes to improve our current classification system.

\subsection{Characterization and Understanding of Complex Diseases -- Big Data Science Initiative (BDSI) with the Department of Veterans Affairs (VA)}

In early 2016 we started a partnership with the Department of Veterans Affairs (VA), which we call MVP-CHAMPION (Computational Health Analytics for Medical Precision to Improve Outcomes Now). As the program has grown both in data and vision, it has grown into a Big Data Science Initiative (BDSI). The aims of the program are to improve the health and well-being of the veterans, and the population as a whole, through better understanding of underlying causes of diseases and conditions, hereditary factors and health history while driving AI, high-performance computing, and data science convergence. We are focusing on studying underlying hereditary, lifestyle, and demographic factors of the diseases such as prostate and liver cancer, clinical depression (leading to suicide), and heart disease. The program uses two of the largest health data assets, in the US and in the world, respectively -- Million Veterans Program (MVP) genomic databank \cite{gaziano2016million}, which is one of the largest global genomic datasets, and VA's Corporate Data Warehouse (CDW) \cite{perrin2010vha,begoli2016towards}, which houses the health data of more than 22 million veterans, and is one of the most complete health datasets. CDW is the primary source for all business intelligence, analytics, and health services research in the VA. As of 2016, it was storing data for some 22 million patients including 7.7 billion lab results, 4.5 billion clinical orders, 3.2 billion clinical notes, 1.4 billion appointments, 2.4 billion encounters, 1.4 million surgeries, 1.3 oncological treatments, 202 million radiology procedures,  71 million immunizations, 2.2 billion pharmacy files, 2.2 billion health factors, 3.3 billion vital signs, 17 million admissions, and 315 million consults. 

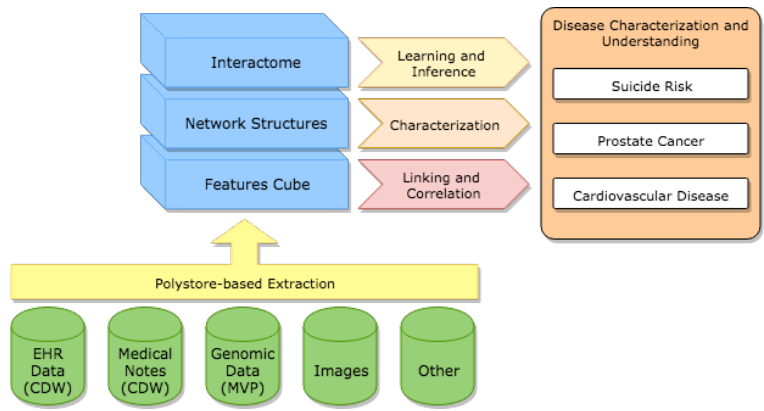
\fig{width=10cm}{heterogeneous_study}{Heterogeneous data is translated into computing-optimized structures that serve as a foundation for downstream analytic and AI processing aimed at diseases characterization and understanding.}

The richness of the data (\figref{heterogeneous_study}) and the complexities of the challenges faced by the VA have created partnerships that drive both next generation technologies and our path to ~\figref{Cog_engine} (c) and advance VA precision medicine priorities. The topics include future high-impact opportunities, crosscutting technologies, and can be summarized as follows:

\noindent \underline{Markedly enhanced prediction and diagnosis of Cardiovascular Disease (CVD)} CVD is the leading cause of death in US men and women – including Veterans – and the cost of care for CVD conditions is high. A collaborative CVD project would build predictive tools that (1) identify improved sets of risk factors for specific types of CVD, and (2) develop methods to inform individualized drug therapies to prevent, preempt and treat CVD. The new tools will enhance prediction, diagnosis and management of major CVD subtypes in Veterans. 

\noindent \underline{Precision discrimination of lethal from non-lethal Prostate Cancer} Many prostate cancer patients undergo surgery or other treatments with significant side effects without knowing whether the risk of such treatments outweigh the benefits, since \textit{a priori} data are limited for many patients regarding lethality. The collaborative prostate cancer project will build improved classifiers for prostate cancer that may significantly aid health care providers in distinguishing lethal from non-lethal prostate cancers. Reducing unnecessary treatments will provide an increased quality of life for patients and allow VA to focus resources where most effective. 

\noindent \underline{Patient-specific analysis for Suicide Prevention} Suicide is the 10th leading cause of death in the US, and is significantly higher in the Veteran population, accounting for 20-22 deaths per day \cite{lyon2017new}. Efforts would improve identification of patients at risk for suicide through new patient-specific algorithms built to provide tailored and dynamic suicide risk scores for each veteran at risk. Working closely with VA’s Office of Suicide Prevention, the tools would be used to create a clinical decision support system that assists VA clinicians in suicide prevention efforts, and helps to evaluate effectiveness of various prevention strategies. 

\noindent \underline{Crosscutting Advances in methods and technology}  Cutting across the three projects are analytic methods that can be scaled to apply to research using the VA EHR and MVP data, as well as requirements for next generation AI and BD analytics. The crosscutting methods include advanced methods in genomic imputation to harness genotype and sequence data in MVP; phenome-wide association studies (PheWAS) \cite{denny2010phewas}, and methods in pharmacogenetics \cite{roses2000pharmacogenetics} analysis. The analytics requirements push development of DOE technologies in key areas including large-scale data analytics, computer modeling, large-scale machine learning, information extraction, and natural language processing as well as algorithms for cost-sensitive decision making under uncertainty. Success in developing these enabling technologies will have large impacts on DOE missions including science, energy, and national security.

In addition to these efforts, there are additional future high impact opportunities. This includes: Predict and control effects of multiple drugs: \textit{polypharmacy}, defined as use of five or more medications at a time, may be our country’s number one drug problem, and it is associated with non-adherence, adverse drug events, falls, inappropriate prescribing, hospitalization, and mortality \cite{rambhade2012survey}; Computer interpretable clinical notes: An estimated 80\% of all electronic medical record information is stored only in unstructured data \cite{murdoch2013inevitable}, including clinical notes by healthcare providers and clinical reports. Here natural language processing (NLP) and information extraction (IE) technologies are being developed to extract information from unstructured (medical) records, allowing it to be combined with other medical data to accelerate understanding and improve patient outcomes; Knowledge enabling classifiers for disease: Disease is often grouped into general categories that define a constellation of symptoms, the organ system affected, or some other outward manifestation. 
 
\subsection{AI/HPC-based Drug Discovery -- Accelerating Therapeutics for Opportunities in Medicine (ATOM) Partnership}

The process of drug discovery, as currently in place, is a long, and costly process, and with a high rate of failure \cite{milligan2013model}. Research questions in this often revolve around the question if there is a better way to rapidly get new medicines to patients. Even before human trials begin, we know that the pre-clinical discovery phase of drug development takes an average of 5.5 years, and  absorbs 33 percent of total developmental cost of medicine development. Millions of compounds get tested, thousands are made, and most fail. The phase space of potential small molecule drug candidates is huge - on the order of $10^{60}$ potential molecules - making the exhaustive, traditional discovery process almost impossible. Consequently, judicious choices in navigating this complex design space must be made. 
Clinical success rates are still only 12\%, indicating poor translation to human patients. 
There is a critical need to accelerate the development of more effective therapies, and to make the process more efficient. To this end, we are partnering with a pharmaceutical \& technology companies, medical research laboratories and institutes to develop a new, computationally enabled starting point -- transforming drug discovery from a slow, sequential, and high-failure process into a rapid, integrated, and patient-centric model driven by high-performance computing.
The ATOM partnership between GlaxoSmithKline, the NCI, DOE, and UC San Francisco - now expanding to include new partners - aims to integrate shared access to previously unused pharmacy data sets, the unique AI/HPC/BD technologies, and new approaches to characterize biology to implement new ways to get medicines to patients, going from target to first in human experimental trials in twelve months or less. The initial efforts are cancer-focused.  

ATOM utilizes a laboratory composed of leading technologies, physiologically relevant complex biological models, and pre-competitive data from GSK on millions of compounds that have been biologically evaluated with access to trillions of compounds in their libraries. It combines the global research and development leadership of the parent organizations with the agile innovative culture of a startup to bring new medicines to the patient. It enables access to new models generated through machine learning, simulation and experiments including the validation of experiments that would accelerate pharmaceutical candidate identification from one year to a few months; and it promotes interdisciplinary work with nontraditional partners and increased interactions between communities that may not often intersect, for example, biochemists, biologists, chemists, molecular biologists, computational scientists, infectious diseases specialists, computing engineers, and the oncology community.

\fig{width=10cm}{ATOM}{Accelerating Therapeutics for Opportunities in Medicine (ATOM)}

\subsection{International: US-Norway Partnership and Cancer Risk Factors}
The Cancer Registry of Norway (CRN) \cite{larsen2009data} is the oldest, and one of the most complete cancer registries in the world, sourced from the single-payer healthcare system, and from the well-understood population. The partnership with CRN provides an opportunity for us to test tools on this unique data asset. The initial efforts are focused on predictive models of cervical cancer screening sequences with attention on optimization of testing frequency. It uses patient questionnaires and HPV status as covariates. The collaboration is currently advancing from 100,000 to over one million cases. There is an ongoing discussions with other Nordic countries about joining efforts to co-develop tools and to co-locate researchers. The initial effort, jointly with the Oslo Cancer Cluster \cite{isaksen2014emergence}, considers a database of 1,728,336 unique Norwegian women’s cervical cancer screening results with multiple screening results are recorded over time for each patient. The data covers a 25-year period, $1991-2015$, with screening times unique to each patient. Overall, there are over ten million individual records.

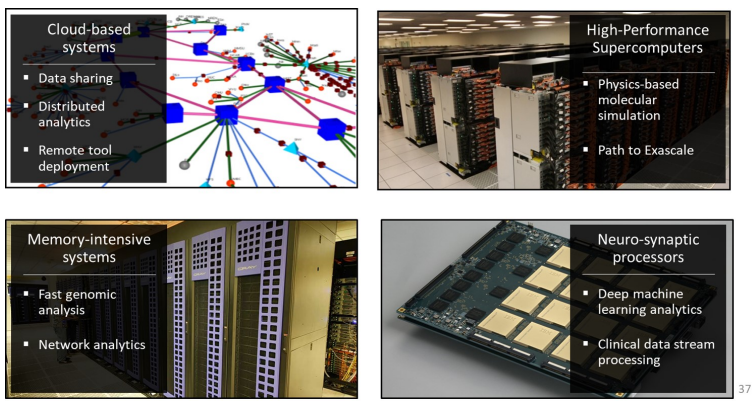
\fig{width=10cm}{convergence2}{We are using precision medicine data as an accelerator for next generation technologies in high performance computing}

\section{Summary}

The demands of UQ in computer prediction, a problem we believe to be NP-Hard, cannot be met on our current HPC technology path. We see that cognitive computing, defined through the technology convergence of AI, Big Data and HPC is an essential next step. With vendor technology decisions being made now and in the next few years in AI and HPC, it is urgent that broad classes of HW and SW are explored to best leverage commercial technology roadmaps. To that end, we are using precision medicine data as a force multiplier and accelerator. This rich, complex, unstructured, heterogeneous, curated, massive data is likely the richest class of data to work on today and brings with it unique partnerships that buys down risk in exploring the many splintered paths forward each with their own tough challenges and also shares costs. We are using this to inform architectures that can integrate neuromorphic processors, GPGPUs, FPGAs, spiking neurons, or other innovations into HPC so we can reach the type of convergence discussed in Section 2 and~\figref{Cog_engine} as part of HW/SW path forward. It is also a problem set that provides a clean-slate: it is not burdened by how we believe we should solve the problem. We are already modifying our path to next generation exascale systems based on the work outlined above. We will need not only advancements in mathematical frameworks for UQ that work with AI and ML, but we will need to understand how we bring together learned models with predictive models into a common framework. As we build new architectures, technologies and tools based on the
complexity of the data, we have an opportunity to push the frontier of precision medicine and next
generation high-performance computing simultaneously. New ideas and broadening the conversations and partnerships are welcome and an important part
of making any progress.

\ack{ } 

\openaccess


\bibliographystyle{superfri}
\bibliography{references}

\begin{thebibliography}{10}
\providecommand{\url}[1]{\texttt{#1}}
\providecommand{\urlprefix}{URL }

\bibitem{begoli2016towards}
Begoli, E., Kistler, D., Bates, J.: Towards a heterogeneous, polystore-like
  data architecture for the us department of veteran affairs (va) enterprise
  analytics. In: 2016 IEEE International Conference on Big Data (Big Data). pp.
  2550--2554 (Dec 2016)

\bibitem{cubitt2012extracting}
Cubitt, T.S., Eisert, J., Wolf, M.M.: Extracting dynamical equations from
  experimental data is np hard. Physical review letters  108(12),  120503
  (2012)

\bibitem{denny2010phewas}
Denny, J.C., Ritchie, M.D., Basford, M.A., Pulley, J.M., Bastarache, L.,
  Brown-Gentry, K., Wang, D., Masys, D.R., Roden, D.M., Crawford, D.C.: Phewas:
  demonstrating the feasibility of a phenome-wide scan to discover
  gene--disease associations. Bioinformatics  26(9),  1205--1210 (2010)

\bibitem{gao2017hierarchical}
Gao, S., Young, M.T., Qiu, J.X., Yoon, H.J., Christian, J.B., Fearn, P.A.,
  Tourassi, G.D., Ramanthan, A.: Hierarchical attention networks for
  information extraction from cancer pathology reports. Journal of the American
  Medical Informatics Association  (2017)

\bibitem{gaziano2016million}
Gaziano, J.M., Concato, J., Brophy, M., Fiore, L., Pyarajan, S., Breeling, J.,
  Whitbourne, S., Deen, J., Shannon, C., Humphries, D., et~al.: Million veteran
  program: a mega-biobank to study genetic influences on health and disease.
  Journal of clinical epidemiology  70,  214--223 (2016)

\bibitem{gligorijevic2016integrative}
Gligorijevic, V., Malod-Dognin, N., Przulj, N.: Integrative methods for
  analyzing big data in precision medicine. Proteomics  16(5),  741--758 (2016)

\bibitem{haring2015traumatic}
Haring, R.S., Narang, K., Canner, J.K., Asemota, A.O., George, B.P.,
  Selvarajah, S., Haider, A.H., Schneider, E.B.: Traumatic brain injury in the
  elderly: morbidity and mortality trends and risk factors. Journal of surgical
  research  195(1),  1--9 (2015)

\bibitem{hawgood2015precision}
Hawgood, S., Hook-Barnard, I., O'Brien, T., Yamamoto, K.: Precision
  medicine:beyond the inflection point. Science Translational Medicine  7(300)
  (2015)

\bibitem{isaksen2014emergence}
Isaksen, A.: Emergence of clusters: by chance or by design. the rise of the
  oslo cancer cluster  (2014)

\bibitem{twoclass}
Kusnezov, D.: How big can you think? challenges at the frontier. Computing in
  Science \& Engineering  9,  62--67 (2007)

\bibitem{larsen2009data}
Larsen, I.K., Sm{\aa}stuen, M., Johannesen, T.B., Langmark, F., Parkin, D.M.,
  Bray, F., M{\o}ller, B.: Data quality at the cancer registry of norway: an
  overview of comparability, completeness, validity and timeliness. European
  journal of cancer  45(7),  1218--1231 (2009)

\bibitem{lyon2017new}
Lyon, J.: New data on suicide risk among military veterans. Jama  318(16),
  1531--1531 (2017)

\bibitem{manley2013traumatic}
Manley, G.T., Maas, A.I.: Traumatic brain injury: an international
  knowledge-based approach. Jama  310(5),  473--474 (2013)

\bibitem{milligan2013model}
Milligan, P., Brown, M., Marchant, B., Martin, S., Graaf, P., Benson, N.,
  Nucci, G., Nichols, D., Boyd, R., Mandema, J., et~al.: Model-based drug
  development: a rational approach to efficiently accelerate drug development.
  Clinical Pharmacology \& Therapeutics  93(6),  502--514 (2013)

\bibitem{murdoch2013inevitable}
Murdoch, T.B., Detsky, A.S.: The inevitable application of big data to health
  care. Jama  309(13),  1351--1352 (2013)

\bibitem{vandv}
Oberkampf, W., Roy, C.: Verification and Validation in Scientific Computing.
  Cambridge University Press (2010)

\bibitem{perrin2010vha}
Perrin, R.A., Bollinger, M.J.: Vha corporate data warehouse height and weight
  data: opportunities and challenges for health services research. Journal of
  rehabilitation research and development  47(8),  739 (2010)

\bibitem{qiu2018deep}
Qiu, J.X., Yoon, H.J., Fearn, P.A., Tourassi, G.D.: Deep learning for automated
  extraction of primary sites from cancer pathology reports. IEEE journal of
  biomedical and health informatics  22(1),  244--251 (2018)

\bibitem{rambhade2012survey}
Rambhade, S., Chakarborty, A., Shrivastava, A., Patil, U.K., Rambhade, A.: A
  survey on polypharmacy and use of inappropriate medications. Toxicology
  international  19(1), ~68 (2012)

\bibitem{ries2003cancer}
Ries, L.A.G., Reichman, M.E., Lewis, D.R., Hankey, B.F., Edwards, B.K.: Cancer
  survival and incidence from the surveillance, epidemiology, and end results
  (seer) program. The oncologist  8(6),  541--552 (2003)

\bibitem{roses2000pharmacogenetics}
Roses, A.D.: Pharmacogenetics and the practice of medicine. Nature  405(6788),
  857 (2000)

\bibitem{yoon2016multi}
Yoon, H.J., Ramanathan, A., Tourassi, G.: Multi-task deep neural networks for
  automated extraction of primary site and laterality information from cancer
  pathology reports. In: INNS Conference on Big Data. pp. 195--204. Springer
  (2016)

\bibitem{yue2013towards}
Yue, J., Vassar, M., Lingsma, H., Cooper, S., Okonkwo, D., Valadka, A., Gordon,
  W., Maas, A., Mukherjee, P., Yuh, E., Puccio, A., Schnyer, D., Manley, G.,
  et~al.: Transforming research and clinical knowledge in traumatic brain
  injury pilot: Multicenter implementation of the common data elements for
  traumatic brain injury. Journal of Neurotrauma  30(22),  1831--1844 (2013)

\end{thebibliography}


\end{document}